\documentclass[a4paper]{article}
\usepackage{xcolor}
\usepackage{enumitem}

\usepackage{INTERSPEECH2018}

\title{A Study of Enhancement, Augmentation, and Autoencoder Methods\\ for Domain Adaptation in Distant Speech Recognition}

\name{Hao Tang, Wei-Ning Hsu, Fran\c cois Grondin, James Glass}
\address{Computer Science and Artificial Intelligence Laboratory\\
Massachusetts Institute of Technology\\
    Cambridge, MA 02139, USA}
\email{\texttt{\{haotang,wnhsu,fgrondin,glass\}@mit.edu}}

\begin{document}

\maketitle

\begin{abstract}
Speech recognizers trained on close-talking speech do not
generalize to distant speech and the word error rate degradation
can be as large as 40\% absolute.  Most studies focus on tackling
distant speech recognition as a separate problem, leaving
little effort to adapting close-talking speech recognizers
to distant speech.  In this work, we review several approaches
from a domain adaptation perspective.
These approaches, including speech enhancement, multi-condition training,
data augmentation, and autoencoders, all involve a transformation
of the data between domains.
We conduct experiments on the AMI data set, where these
approaches can be realized under the same controlled setting.
These approaches lead to different amounts of improvement
under their respective assumptions.
The purpose of this paper is to quantify and characterize the 
performance gap between the two domains, setting up the basis for studying
adaptation of speech recognizers from close-talking speech to distant speech.
Our results also have implications for improving distant speech recognition.
\end{abstract}

\section{Introduction}

Domain adaptation refers to the task of adapting models
trained on one domain to other domains.
In the general setting,
models are trained in the source domain and tested on the target domain.
The source domain may or may not have overlaps with the target domain.
The mismatch between the training and the test conditions
causes the task performance to deteriorate, because
generalization guarantees rely on the assumption that
the training and test samples come from the same underlying distribution.
Domain adaption in the most general case is possible
under some assumptions, but deemed challenging \cite{B+2010,G+2016}.

Domain adaptation for speech recognition is particularly
difficult considering the mismatch in speakers, speaking styles, noise types, and room acoustics etc.
There has been significant success in dealing with speaker mismatch,
for example, adapting a speaker-independent model
to a known speaker or even adapting to an unknown speaker \cite{GL1994,LW1995}.
Developing speech recognizers that are robust to many noise types is
more challenging, and in theory it is impossible to have
a model that is robust to any adversarial noise \cite{BLLP2010}. 
It is, however, still possible to design speech recognizers that are
robust to natural noise types that occur in our daily lives.
Significant progress has been made in this direction, especially
when the noise types are known at training time,
for example, with speech enhancement techniques or
multi-condition training \cite{SYW2013,Y+2012,W+2015}.

This paper focuses on the task of adapting
speech recognizers trained on close-talking speech to distant speech.
Distant speech recognition is itself a difficult
task \cite{WM2009}.
The difficulty is often attributed to reverberation, i.e., weaker copies
of the original speech signals. Early reverberation is
considered easy to handle, because convolving shifted
impulses in the time domain is nothing but a constant scaling
function on the power spectrum.
Late reverberation, on the other hand, is not limited to
single short-term spectra and cannot be approximated
well with shifted impulses. As a result, the speech
is corrupted with a type of noise that is highly correlated
with the speech from the past.
Important effort has been devoted to training models directly
on distant speech \cite{SGR2013}.
Other solutions for distant speech recognition include
using multiple microphones \cite{D+2016, HDH2016, E+2016}, 
speech enhancement techniques \cite{D+2014,W+2015},
and data augmentation \cite{KPPSK2017,KPPK2015}.
It is also unclear if the degradation in performance
is really due to reverberation and not due to other causes,
such as the difference in gain levels.
We investigate this by training
models on data augmented with simulated reverberation.

There has been some work in adapting speech recognizers to
distant speech \cite{SGR2013,H+2015,QTY2016,QTYZ2016}. However, different
studies use different settings, for example, whether
it is allowable to use parallel data to train models,
or whether we have access to labels in the other domain.
In this paper, we consider various settings, their requirements,
and the performance of speech recognizers of a fixed
architecture.  The purpose of this paper is to quantify
and characterize the gap of these settings, and set up
the basis for studying domain adaptation for distant
speech recognition.  Note that we do not consider the
online adaptive setting, a common scenario for speaker adaptation \cite{GL1994,LW1995,N+1995,LS2010},
where we have a small amount of labeled data to adapt to the target domain.


\section{Domain Adaptation}

In this section, we summarize the approaches and their requirements
for adapting speech recognizers from close-talking speech to
distant speech.
In the general setting, let the input space be $\mathcal{X}$
and the output space be $\mathcal{Y}$.
For speech recognition, $\mathcal{X}$ is the set of sequences of log Mel filterbank feature vectors,
and $\mathcal{Y}$ is the set of word sequences.
We have two unknown data distributions
$\mathcal{D}_1$ and $\mathcal{D}_2$ over $\mathcal{X} \times \mathcal{Y}$ representing the
source and the target domain.
In the following discussion,
we refer to close-talking speech as the source domain
and distant speech as the target domain.

\subsection{Speech enhancement}

To reduce the mismatch between domains,
a simple approach is to transform data from the target domain
to the source domain where the recognizer is trained.
We assume there is an unknown distortion function $C: \mathcal{X} \to \mathcal{X}$ such that $C(x) \sim \mathcal{D}_2$ for $x \sim \mathcal{D}_1$.
The goal is to find a function $T: \mathcal{X} \to \mathcal{X}$ such that $T(C(x)) \approx x$ for $x \sim \mathcal{D}_1$.
For speech processing, transforming noisy speech to clean
speech is referred to as speech enhancement.

In general, speech enhancement has a broader goal: transforming signals
so that the speech stands out and becomes more audible.
This typically involves removing noise (though sometimes adding
noise can improve intelligibility~\cite{WHBBW1997}).
We focus on the limited sense of enhancement,
making speech closer to its clean counterpart
while ignoring intelligibility.
We assume we have access to a parallel data set $\{(x_1, \tilde{x}_1), \dots, (x_n, \tilde{x}_n)\}$ where $\tilde{x}_i = C(x_i)$ for $i = 1, \dots, n$.
The objective is to approximate the clean speech $x_i$
given the noisy speech $\tilde{x}_i$ by
minimizing the Euclidean distance
$\|x_i - T(\tilde{x}_i)\|^2_2$ for $i=1, \dots, n$.
Minimizing this objective for speech enhancement
was first proposed in \cite{EM1984}
and is explored in the context of neural networks in \cite{WN1998}.
Deep neural networks are particularly suitable for speech
enhancement without posing any assumptions on the noise types.
Modern treatments with deep networks are studied
in \cite{M+2012,SYW2013,XDDL2014,W+2015}.

Once a model for speech enhancement is trained,
we enhance the speech signal prior to doing speech recognition,
i.e., using $T(\tilde{x})$ instead of $\tilde{x}$ as the input
to the speech recognizer.
Training speech enhancement models requires
parallel data in both domains, which makes
data collection costly.
However, this approach does not need transcripts for the parallel
data.  Speech recognizers trained on the source domain
can also be reused without additional training.

\subsection{Multi-condition training}

Another simple approach to reduce the mismatch between domains
is to use the data from the target domain during training.
Suppose we have $S_1 \sim \mathcal{D}_1^n$ and $S_2 \sim \mathcal{D}_2^m$ where $n$ and $m$ are the numbers of samples
for the two data sets.
Models are simply on the data set $S_1 \cup S_2$.
If the performance on the target domain is the only concern,
we can always discard $S_1$ and train models only on $S_2$.
For noise-robust speech recognition, training models
on different noise conditions is referred to as
multi-condition training or multi-style training.
Multi-condition training can be traced back to \cite{PM1988},
and has been shown to reduce mismatch for different noise conditions \cite{HES2000}.
Deep neural networks work particularly well with
multi-condition training due to the large model capacity \cite{Y+2012,SGR2013}.

Multi-condition training requires labeled data in both domains,
so data collection can be costly.
Additional training, either from scratch or from a pre-trained
model, is required.
When the model capacity is large enough, a single model is
able to cover multiple domains.
However, the training time scales linearly with the
amount of data.

\subsection{Data augmentation}

As a special case of multi-condition training,
data augmentation transforms
data from the source domain to the target domain
(i.e., the opposite of speech enhancement).
This typically involves corrupting the clean data
with different noise types or transforming the clean
data with simulators, such as convolving the clean
speech with room impulse responses.
Formally, we assume we have a generator distribution $G(\hat{x}|x)$.
Let
$G(S) = \{(\hat{x}_1, y_1), \dots, (\hat{x}_n, y_n)\}$
where $\hat{x}_i \sim G(\hat{x}|x_i)$ for $i=1, \dots, n$ and
some data set $S = \{(x_1, y_1), \dots, (x_n, y_n)\} \sim \mathcal{D}_1^n$.
We train models on the data set $S \cup G(S)$.
This approach is expected to work well
if the generator is able to the match the target domain,
i.e., for $x \sim \mathcal{D}_1$ and $\hat{x} \sim G(\hat{x}|x)$,
either $\hat{x} \sim \mathcal{D}_2$ or
$\hat{x} \approx C(x)$ for
an unknown distortion function $C$ such that
$C(x) \sim \mathcal{D}_2$.
Data augmentation was originally designed as a regularization technique
for learning transformation invariant
features, and has been successful in
image classification tasks with convolutional neural networks \cite{LBBH1998, SSP2003, KSH2012}.
Data augmentation has been applied to speech
recognition in \cite{JH2013, KPPK2015, KPPSK2017}.

Data augmentation is suitable when the simulation of
noise or other factors is simple, for example,
perturbing vocal tract lengths \cite{JH2013},
perturbing speed \cite{KPPK2015}, and simulating
reverberation \cite{KPPSK2017}.
Data from the target domain is not required.
However, the training time scales linearly with
the amount of generated data.

\subsection{Unsupervised domain adaptation with autoencoders}

Finding similarities between the target and the source
domains is yet another way to tackle domain mismatch.
For example, we assume a common distribution for
linguistic content, such as English utterances.
The source and the target domain
can still have their own nuisance factors depending on speakers
and channels.  Each domain can be modeled as a generative
process where an utterance is first sampled from
the shared distribution and is transformed according 
to the nuisance factors.
Since the two domains are symmetric, we describe
the process in one domain $\alpha$; the other domain follows
the same generative story.
For example, we can have $\alpha \in \{0, 1\}$ where 0 denotes the
source domain and 1 denotes the target domain.
Suppose an utterance from domain $\alpha$ has $K$ segments $s_1, \dots, s_K$.
Each segment $s_k$ is generated by
a domain-independent vector $z_k^1$
and a domain-dependent vector $z_{k, \alpha}^2$.
The domain-independent vector $z_k^1 \sim \mathcal{D}$
encodes the linguistic content where $\mathcal{D}$ is
the shared distribution for all domains,
while the domain-independent vector
$z_{k, \alpha}^2 \sim \mathcal{D}_\alpha$
encodes the nuisance factors,
such as speakers and channels, specific to domain $\alpha$.
The segment $s_k$ is then generated from
a function that depends on $z_k^1$ and $z_{k, \alpha}^2$

We use factorized hierarchical variational autoencoders (FHVAE) \cite{HG2017}
to model the above generative process
with two inference networks $q(z^2|x)$, $q(z^1|x, z^2)$.
Without any further constraints, $z^1$ and $z^2$ are fully exchangeable.
To make sure $z^2$ captures the nuisance factors,
we constrain the $z^2$'s from the same utterance
to be similar while leaving $z^1$ unconstrained,
because the nuisance factors largely remain
unchanged within the same utterance.
In addition, there is a loss enforcing $z^2$ to be predictive
of the utterance identity.

After training the FHVAE on all data combined, we use the inference network
to obtain the vectors that encode the linguistic contents
and discard the vectors for the nuisance factors.
Speech recognizers are trained on these new set of features.
This approach does not require parallel data from both domains,
and the data does not need to be labeled.
However, tuning FHVAEs might be difficult.
If the model has too many parameters for reconstruction,
we might obtain a trivial identity function.
If the weight between reconstruction and the KL-divergence
is tuned, we do not have a fixed objective to compare different FHVAEs.

\section{Experiments}

In order to have a fair comparison for
all the settings,
we conduct experiments on the AMI data set,
where parallel recordings and labels are available for
both the close-talking and the distant speech domains.
The AMI data set is
a meeting corpus with around 100 hours of conversational
non-native English speech.  The meetings are recorded in a controlled environment
with independent headset microphones (IHM) on each speaker
and multiple distant microphones.  The audio streams from different
microphones are aligned with beamforming.
We take the aligned recordings from the IHMs and one specific distant microphone,
referred to as the single distant microphone (SDM), for
our experiments.
To avoid excessively querying the standard test set,
we do not report numbers on the standard test set.
Instead, we use 90\% of the training set for training, leave 10\%
for development, such as step size tuning and early stopping,
and only report word error rates (WERs) on the standard development set.

Following \cite{Z+2015}, we use 80-dimensional log Mel
filterbank features, and train two speaker-adaptive
hidden Markov models (HMM), one for IHM and one for SDM.
We obtain forced alignments of the tied HMM states
(also known as pdf-ids) for both IHM and SDM
recordings with the corresponding systems, and use the pdf-ids as targets for acoustic model training.
We use eight-layer time-delay neural networks (TDNNs)
with 1000 hidden hidden units per layer as our acoustic models.
Following~\cite{PWPK2018}, the context sizes of
the TDNNs from layer one to seven are
$[-1, 0, 1], [-1, 0, 1], [-1, 0, 1], [-3, 0, 3], [-3, 0, 3], [-3, 0, 3]$,
where $[i, 0, k]$ indicates the summation of hidden vectors
at time $t + i$, $t$, and $t + k$.  Formally,
to compute the hidden layer $h_{\ell}$ from $h_{\ell - 1}$
with context $[i, 0, k]$, we have
\begin{align}
\tilde{h}_t & = W_\ell h_{\ell - 1, t} + b_\ell \\
h_{\ell, t} & = \text{ReLU}(\tilde{h}_{t + i} + \tilde{h}_t + \tilde{h}_{t + k})
\end{align}
Note that in contrast to the standard recipe, we only use
the 80-dimensional log Mel features as input without appending
the i-vectors.

\subsection{Baseline and multi-condition training}

Since we are interested in adapting models trained on close-talking
speech to distant speech, we train two TDNNs, one on IHM and
one on SDM, and test them on utterances from both IHM and SDM.
We use stochastic gradient descent (SGD) with a fixed step size 0.025
and a mini-batch size of 1 utterance
to optimize the cross entropy for 20 epochs.
Gradients are clipped to norm 5.
We choose the best performing model from the 20 epochs based on the
frame error rates, and train it for another 5 epochs
with step size $0.025 \times 0.75$ decayed by 0.75
after each epoch.
Results are shown in Table~\ref{tbl:baseline}.
The WER increases from 27.4\% to 70.3\% when using a close-talking model
on distant speech.
Note that in the SDM setting, the WER of IHM is lower
than the that of SDM, consistent with the results reported in \cite{H+2015,PMWPK2016}.
We also confirm the improvement reported
in \cite{PMWPK2016}, training models on SDM data while
using IHM alignments.
For consistency, we use IHM alignments for the rest of the experiments.

For multi-condition training, we train and tune TDNNs on both
IHM and SDM combined.  The training procedure remains the same
except that we use a smaller initial step size 0.01.
Results are shown in Table~\ref{tbl:baseline}.
The TDNN is able to match the results on both domains.

\begin{table}
\begin{center}
\caption{\label{tbl:baseline} WERs (\%) for models trained and tested on various domains.}
\begin{tabular}{llll}
train & target  &        \\
\hline
IHM   & IHM     & 27.4   \\
IHM   & SDM     & 70.3   \\
\hline
SDM   & IHM     & 41.8   \\
SDM   & SDM     & 49.7   \\
SDM (IHM alignments) 
      & IHM     & 39.2   \\
SDM (IHM alignments) 
      & SDM     & 46.6   \\
\hline
IHM + SDM
      & IHM     & 27.2  \\
IHM + SDM
      & SDM     & 45.3
\end{tabular}
\end{center}
\end{table}

\subsection{Data augmentation with simulated reverberation}

To investigate the impact of reverberation on distant speech
recognition, we use the image method described in \cite{AB1979} to create a set of
simulated room impulse responses (RIRs) with different
rectangular room sizes, speaker positions, and microphone
positions, as proposed in \cite{KPPSK2017}.
Three sets of rooms ($\mathcal{S}_{1}$, $\mathcal{S}_{2}$, and $\mathcal{S}_{3}$) are generated by uniformly sampling the width $L_{x}$, length $L_{y}$ and height $L_{z}$ (in meters) in set-wise ranges (where $\mathcal{U}(a,b)$ stands for a uniform distribution between $a$ and $b$): 
\begin{align*}
\mathcal{S}_{1} &: L_{x} \sim \mathcal{U}(1,10), L_{y} \sim \mathcal{U}(1,10), L_{z} \sim \mathcal{U}(2,5) \\
\mathcal{S}_{2} &: L_{x} \sim \mathcal{U}(10,30), L_{y} \sim \mathcal{U}(10,30), L_{z} \sim \mathcal{U}(2,5) \\
\mathcal{S}_{3} &: L_{x} \sim \mathcal{U}(30,50), L_{y} \sim \mathcal{U}(30,50), L_{z} \sim \mathcal{U}(2,5)
\end{align*}
For each room, the speed of sound is set to $343$ m/sec, and the wall, ceiling and floor reflection coefficient is sampled from a uniform distribution between $0.2$ and $0.8$.
For each set, 200 rooms are sampled, 100 RIRs are obtained for both the source
and microphone positioned randomly in the room, leading to a total of 60,000 simulated RIRs for all sets.
For each utterance in the dataset, one of these RIRs is selected randomly and convolved with the clean speech.

We train TDNNs on IHM and IHM corrupted with simulated reverberation
the same way we train multi-conditioned models in the previous
section.  We test the TDNNs on the IHM data corrupted with
reverberation and SDM.  Results are shown in Table~\ref{tbl:reverb}.
The degradation due to reverberation is not as severe compared
to that of SDM.  Training TDNNs with the additional data
does help generalize to the SDM domain.  However, the improvement
is far from closing the gap, suggesting that
reverberation might not be the major cause of the performance
degradation.

\begin{table}
\begin{center}
\caption{\label{tbl:reverb} WERs (\%) for models trained
with data augmentation and tested on various domains, where IHM-r denote
the domain with data corrupted with simulated reverberation.}
\begin{tabular}{lll}
train & target   &        \\
\hline
IHM   & IHM      & 27.4   \\
IHM + IHM-r & IHM      & 28.7  \\
\hline
IHM   & IHM-r    & 59.3  \\
IHM + IHM-r & IHM-r    & 43.7  \\
\hline
IHM   & SDM      & 70.3   \\
IHM + IHM-r & SDM      & 63.3  
\end{tabular}
\end{center}
\end{table}

\subsection{Speech enhancement}

For speech enhancement, we use the same TDNN architecture
without the final softmax.  TDNNs are trained to
predict the features of IHM utterances given the corresponding
features of SDM utterances, while being an identity function
given features of IHM utterances.
The training set in this case is the IHM and SDM combined.
The mean squared error is minimized using the same training procedure
with an initial step size of 0.01.
After training the enhancement model, we take the baseline model
trained on IHM and test it on the enhanced data.
Results are shown in Table~\ref{tbl:enhancement}.
The WER on the enhanced SDM (SDM-e) is significantly reduced
from 70.3\% to 54.2\%, while maintaining the WER on the IHM domain.

Again, to investigate how reverberation plays a role in
distant speech recognition, we train a dereveberation
TDNN on the IHM data corrupted with reverberation while
being an identity function on the clean IHM data.
We then evaluate the baseline TDNN trained on IHM
with the dereveberated data.  Results are shown in
Table~\ref{tbl:enhancement}.  We see some amount of
improvement from 59.3\% (in Table~\ref{tbl:reverb}) to 53.1\%,
suggesting that the TDNN is able to perform blind dereverberation.
However, the improvement is not as large as
the multi-condition TDNN, suggesting that blind
dereverberation is in itself a challenging task.
We also evaluate the
dereverberation model on SDM, and find no improvement
over the baseline.  This again suggests that the domain
mismatch between IHM and SDM might not be due to reverberation but
some other types of mismatch.

\begin{table}
\begin{center}
\caption{\label{tbl:enhancement} WERs (\%) for models trained on IHM and tested on various domains,
where IHM-e and SDM-e denote the domains with enhanced data and IHM-dr, IHM-r-dr, and SDM-dr denote the domains with dereveberated data.}
\begin{tabular}{llll}
train & target   &        \\
\hline
IHM   & IHM      & 27.4   \\
IHM   & IHM-e    & 27.8   \\
\hline
IHM   & SDM      & 70.3   \\
IHM   & SDM-e    & 54.2   \\
\hline
IHM   & IHM-dr   & 27.6   \\
IHM   & IHM-r-dr & 53.1   \\
IHM   & SDM-dr   & 70.0   
\end{tabular}
\end{center}
\end{table}

\subsection{Unsupervised domain adaptation with FHVAEs}

For unsupervised domain adaptation,
we train a FHVAE by minimizing
the discriminative segmental variational 
lower bound~\cite{HG2017} with a factor $\alpha=10$
for the utterance discriminative loss.
The FHVAE consists of two encoders and one decoder.
One encoder is for the
shared distribution (representing linguistic content)
and the other is for the domain-specific distribution (representing
nuisance factors).  The decoder takes
the output vectors from both encoders and
reconstructs the input features.
Inputs to an FHVAE are 20 frames of 80-dimensional log Mel features.
Both encoders are LSTM networks~\cite{HS1997} with 256 memory cells that
process one frame at each step, followed by an affine transform layer that 
takes the LSTM output from the last step and predicts the 
posterior mean and log variance of the corresponding latent variables.
We use an LSTM decoder with 256 memory cells, 
where the LSTM output from each step is passed to an affine transform
layer to predict the mean and variance of a frame.
The Adam~\cite{KB2014} optimizer is used with $\beta_1 = 0.95$, 
$\beta_2 = 0.999$, $\epsilon = 10^{-8}$, and initial learning rate of $10^{-3}$.
Early stopping is done by monitoring the evidence lower bound
on the development set.

After the FAVAE is trained, we use the encoder for the shared
distribution to produce features.
A feature vector is generated at each time point
by taking a 20-frame segment centered at the current time point
and feeding it forward into the encoder.
Following~\cite{HG2018}, since the generated feature sequence 
is 19 frame shorter, we repeat
the first and the last feature vector
at each end to match the original length.
The hidden vectors are then normalized by
subtracting the mean and dividing by the standard
deviation computed over the training set.
TDNNs are trained on the produced feature vectors with the same
training procedure as in previous sections.
Since the distribution is modeled as a Gaussian,
we use the Gaussian mean vectors and have the option to include
the log-variance vectors as features.
Results are shown in Table~\ref{tbl:autoenc}.
While there is a small amount of degradation
in the IHM domain, we see an improvement from
70.3\% to 61.8\% in the SDM domain.
This suggests that the SDM features produced
by the FHVAE are closer to IHM in the latent space.
The improvement is even larger than data augmentation
with simulated reverberation.
However, we find that including the log-variance
as features might not help adapting to the target domain.
This needs further investigation.

\begin{table}
\begin{center}
\caption{\label{tbl:autoenc} WERs (\%) for models trained on
hidden vectors produced by an FHVAE, where the rows with $\mu^1$
use the mean of shared distribution and the rows with $\log\sigma^1$
use the log variance of the shared distribution as features.}
\begin{tabular}{llll}
train & target  &        \\
\hline
IHM   & IHM     & 27.4   \\
IHM   & SDM     & 70.3   \\
\hline
IHM-$\mu^1$
      & IHM-$\mu^1$  & 31.8   \\
IHM-$\mu^1$ 
      & SDM-$\mu^1$  & 61.8   \\
\hline
IHM-$(\mu^1, \log\sigma^1)$ 
      & IHM-$(\mu^1, \log\sigma^1)$  & 30.3   \\
IHM-$(\mu^1, \log\sigma^1)$ 
      & SDM-$(\mu^1, \log\sigma^1)$  & 72.9         
\end{tabular}
\end{center}
\end{table}

\section{Conclusion}

In this work, we review several approaches,
including speech enhancement, data augmentation,
and autoencoders, to bridge the gap
from close-talking speech recognition to distant speech
recognition from a domain adaptation perspective.
We find that all approaches are able to produce
models that are more robust than the baseline.
Multi-condition training gives the
best results among all approaches, but it also
has the most stringent requirement, requiring
labeled data in all domains.
Speech enhancement comes second but also has
a stringent requirement, requiring parallel
unlabeled data.  Data augmentation has the potential to match
the performance of multi-condition training.
However, it requires the data generation process
to cover the condition of the target domain.
Unsupervised domain adaptation with autoencoders
is promising, achieving better results than
data augmentation with simulated reverberation
while only requiring independent unlabeled data from
both domains.  Finally, the results suggest that
the mismatch between IHM and SDM in the AMI data set
is probably less about reverberation and has
some other factors, such as cross talking \cite{PMWPK2016}, that need to be
studied further.

\bibliographystyle{IEEEtran}
\bibliography{distant}

\begin{thebibliography}{10}
\providecommand{\url}[1]{#1}
\csname url@samestyle\endcsname
\providecommand{\newblock}{\relax}
\providecommand{\bibinfo}[2]{#2}
\providecommand{\BIBentrySTDinterwordspacing}{\spaceskip=0pt\relax}
\providecommand{\BIBentryALTinterwordstretchfactor}{4}
\providecommand{\BIBentryALTinterwordspacing}{\spaceskip=\fontdimen2\font plus
\BIBentryALTinterwordstretchfactor\fontdimen3\font minus
  \fontdimen4\font\relax}
\providecommand{\BIBforeignlanguage}[2]{{%
\expandafter\ifx\csname l@#1\endcsname\relax
\typeout{** WARNING: IEEEtran.bst: No hyphenation pattern has been}%
\typeout{** loaded for the language `#1'. Using the pattern for}%
\typeout{** the default language instead.}%
\else
\language=\csname l@#1\endcsname
\fi
#2}}
\providecommand{\BIBdecl}{\relax}
\BIBdecl

\bibitem{B+2010}
S.~Den-David, J.~Blitzer, K.~Crammer, A.~Kulesza, F.~Pereira, and J.~W.
  Vaughan, ``A theory of learning from different domains,'' \emph{Machine
  Learning}, vol.~79, 2010.

\bibitem{G+2016}
Y.~Ganin, E.~Ustinova, H.~Ajakan, P.~Germanin, H.~Larochelle, F.~Laviolette,
  M.~Marchand, and V.~Lempitsky, ``Domain-adversarial training of neural
  networks,'' \emph{Journal of Machine Learning}, vol.~17, 2016.

\bibitem{GL1994}
J.-L. Gauvain and C.-H. Lee, ``Maximum a posteriori estimation for multivariate
  {Gaussian} mixture observations of {Markov} chains,'' \emph{IEEE Transactions
  on Acoustics, Speech, and Signal Processing}, 1994.

\bibitem{LW1995}
C.~J. Leggetter and P.~C. Woodland, ``Maximum likelihood linear regression for
  speaker adaptation of continuous density hidden {Markov} models,''
  \emph{Computer Speech \& Language}, vol.~9, 1995.

\bibitem{BLLP2010}
S.~Ben-David, T.~Luu, T.~Lu, and D.~P\'al, ``Impossibility theorems for domain
  adaptation,'' in \emph{International Conference on Artificial Intelligence
  and Statistics}, 2010.

\bibitem{SYW2013}
M.~L. Seltzer, D.~Yu, and Y.~Wang, ``An investigation of deep neural networks
  for noise robust speech recognition,'' in \emph{International Conference on
  Acoustics, Speech and Signal Processing}, 2013.

\bibitem{Y+2012}
T.~Yoshioka, A.~Sehr, M.~Delcroix, K.~Kinoshita, R.~Maas, T.~Nakatani, and
  W.~Kellermann, ``Making machines understand us in reverberant rooms,''
  \emph{IEEE Signal Processing Letter}, 2012.

\bibitem{W+2015}
F.~Weninger, H.~Erdogan, S.~Watanabe, E.~Vincent, J.~L. Roux, J.~R. Hershey,
  and B.~Schuller, ``Speech enhancement with {LSTM} recurrent neural networks
  and its application to noise-robust {ASR},'' in \emph{International
  Conference on Latent Variable Analysis and Single Separation}, 2015.

\bibitem{WM2009}
M.~W\"olfel and J.~McDonough, \emph{Distant speech recognition}.\hskip 1em plus
  0.5em minus 0.4em\relax John Wiley \& Sons, 2009.

\bibitem{SGR2013}
P.~Swietojanski, A.~Ghoshal, and S.~Renals, ``Hybrid acoustic models for
  distant and multichannel large vocabulary speech recognition,'' in \emph{IEEE
  Workshop on Automatic Speech Recognition and Understanding}, 2013.

\bibitem{D+2016}
J.~Du, Y.-H. Tu, L.~Sun, F.~Ma, H.-K. Wang, J.~Pan, C.~Liu, J.-D. Chen, and
  C.-H. Lee, ``The {USTC-iFlytek} system for {CHiME-4} challenge,'' \emph{The
  4th CHiME Speech Separation and Recognition Challenge Workshop}, pp. 36--38,
  2016.

\bibitem{HDH2016}
L.~D. Jahn~Heymann and R.~Haeb-Umbach, ``Wide residual {BLSTM} network with
  discriminative speaker adaptation for robust speech recognition,'' in
  \emph{The 4th CHiME Speech Separation and Recognition Challenge Workshop},
  2016.

\bibitem{E+2016}
H.~Erdogan, T.~Hayashi, J.~R. Hershey, T.~Hori, C.~Hori, W.-N. Hsu, S.~Kim,
  J.~L. Roux, Z.~Meng, and S.~Watanabe, ``Multi-channel speech recognition:
  {LSTM}s all the way through,'' in \emph{The 4th CHiME Speech Separation and
  Recognition Challenge Workshop}, 2016.

\bibitem{D+2014}
J.~Du, Q.~Wang, T.~Gao, Y.~Xu, L.~Dai, and C.-H. Lee, ``Robust speech
  recognition with speech enhanced deep neural networks,'' in
  \emph{Interspeech}, 2014.

\bibitem{KPPSK2017}
T.~Ko, V.~Peddinti, D.~Povey, M.~L. Seltzer, and S.~Khudanpur, ``A study on
  data augmentation of reverberant speech for robust speech recognition,'' in
  \emph{International Conference on Acoustics, Speech and Signal Processing},
  2017.

\bibitem{KPPK2015}
T.~Ko, V.~Peddinti, D.~Povey, and S.~Khudanpur, ``Audio augmentation for speech
  recogntion,'' in \emph{Interspeech}, 2015.

\bibitem{H+2015}
I.~Himawan, P.~Motlicek, D.~Imseng, B.~Potard, N.~Kim, and J.~Lee, ``Learning
  feature mapping using deep neural network bottleneck features for distant
  large vocabulary speech recognition,'' in \emph{International Conference on
  Acoustics, Speech and Signal Processing}, 2015.

\bibitem{QTY2016}
Y.~Qian, T.~Tan, and D.~Yu, ``An investigation into using parallel data for
  far-field speech recognition,'' in \emph{International Conference on
  Acoustics, Speech and Signal Processing}, 2016.

\bibitem{QTYZ2016}
Y.~Qian, T.~Tan, D.~Yu, and Y.~Zhang, ``Integrated adaptation with multi-factor
  joint-learning for far-field speech recognition,'' in \emph{International
  Conference on Acoustics, Speech and Signal Processing}, 2016.

\bibitem{N+1995}
J.~Neto, L.~Almeida, M.~Hochberg, C.~Martins, L.~Nunes, S.~Renals, and
  T.~Robinson, ``Speaker-adaptation for hybrid {HMM-ANN} continuous speech
  recognition system,'' in \emph{European Conference on Speech Communication
  and Technology (EUROSPEECH)}, 1995.

\bibitem{LS2010}
B.~Li and K.~C. Sim, ``Comparison of discriminative input and output
  transformations for speaker adaptation in the hybrid {NN/HMM} systems,'' in
  \emph{Interspeech}, 2010.

\bibitem{WHBBW1997}
R.~M. Warren, K.~R. Hainsworth, B.~S. Brubaker, J.~A. Bashford, and E.~W.
  Healy, ``Spectral restoration of speech: intelligibility is increased by
  inserting noise in spectral gaps,'' \emph{Perception \& Psychophysics},
  vol.~59, 1997.

\bibitem{EM1984}
Y.~Ephraim and D.~Malah, ``Speech enhancement using a minimum mean-square error
  short-time spectral amplitude estimator,'' \emph{IEEE Transactions on
  Acoustics, Speech, and Signal Processing}, vol.~32, 1984.

\bibitem{WN1998}
E.~A. Wan and A.~T. Nelson, ``Networks for speech enhancement,'' in
  \emph{Handbook of Neural Networks for Speech Processing}.\hskip 1em plus
  0.5em minus 0.4em\relax Artech House, 1998.

\bibitem{M+2012}
A.~L. Maas, Q.~V. Le, T.~M. O'Neil, O.~Vinyals, P.~Nguyen, and A.~Y. Ng,
  ``Recurrent neural networks for noise reduction in robust {ASR},'' in
  \emph{Interspeech}, 2012.

\bibitem{XDDL2014}
Y.~Xu, J.~Du, L.-R. Dai, and C.-H. Lee, ``An experimental study on speech
  enhancement based on deep neural networks,'' \emph{IEEE Signal Processing
  Letter}, vol.~21, 2014.

\bibitem{PM1988}
B.~B. Paul and E.~A. Martin, ``Speaker stress-resistant continuous speech
  recognition,'' in \emph{International Conference on Acoustics, Speech and
  Signal Processing}, 1988.

\bibitem{HES2000}
H.~Hermansky, D.~P. Ellis, and S.~Sharma, ``Tandem connectionist feature
  extraction for conventional {HMM} systems,'' in \emph{International
  Conference on Acoustics, Speech and Signal Processing}, 2000.

\bibitem{LBBH1998}
Y.~LeCun, L.~Bottou, Y.~Bengio, and P.~Haffner, ``Gradient-based learning
  applied to document recognition,'' \emph{Proceedings of the IEEE}, 1998.

\bibitem{SSP2003}
P.~Y. Simard, D.~Steinkraus, and J.~C. Platt, ``Best practices for
  convolutional neural networks applied to visual document analysis,'' in
  \emph{International Conference on Document Analysis and Recognition}, 2003.

\bibitem{KSH2012}
A.~Krizhevsky, I.~Sutskever, and G.~E. Hinton, ``Imagenet classification with
  deep convolutional neural networks,'' in \emph{Advances in Neural Information
  Processing Systems}, 2012.

\bibitem{JH2013}
N.~Jaitly and G.~E. Hinton, ``Vocal tract length perturbation ({VTLP}) improves
  speech recognition,'' in \emph{International Conference on Machine Learning},
  2013.

\bibitem{HG2017}
W.-N. Hsu, Y.~Zhang, and J.~Glass, ``Unsupervised learning of disentangled and
  interpretable representations from sequential data,'' in \emph{Advances in
  Neural Information Processing Systems}, 2017.

\bibitem{Z+2015}
Y.~Zhang, G.~Chen, D.~Yu, K.~Yao, S.~Khudanpur, and J.~Glass, ``Highway long
  short-term memory {RNN}s for distant speech recognition,'' in
  \emph{International Conference on Acoustics, Speech and Signal Processing},
  2016.

\bibitem{PWPK2018}
V.~Peddinti, Y.~Wang, D.~Povey, and S.~Khudanpur, ``Low latency acoustic
  modeling using temporal convolution and {LSTM}s,'' \emph{IEEE Signal
  Processing Letters}, vol.~25, 2018.

\bibitem{PMWPK2016}
V.~Peddinti, V.~Manohar, Y.~Wang, D.~Povey, and S.~Khudanpur, ``Far-field {ASR}
  without parallel data,'' in \emph{Interspeech}, 2016.

\bibitem{AB1979}
J.~B. Allen and D.~A. Berkley, ``Image method for efficiently simulating
  small-room acoustics,'' \emph{The Journal of the Acoustical Society of
  America}, vol.~65, no.~4, pp. 943--950, 1979.

\bibitem{HS1997}
S.~Hochreiter and J.~Schmidhuber, ``Long short-term memory,'' \emph{Neural
  computation}, 1997.

\bibitem{KB2014}
D.~P. Kingma and J.~Ba, ``Adam: A method for stochastic optimization,'' in
  \emph{Proceedings of the International Conference on Learning
  Representations}.

\bibitem{HG2018}
W.-N. Hsu and J.~Glass, ``Extracting domain invariant features by unsupervised
  learning for robust automatic speech recognition,'' in \emph{International
  Conference on Acoustics, Speech and Signal Processing}, 2018.

\end{thebibliography}

\end{document}